\documentclass[runningheads]{llncs}
\usepackage{cite}
\usepackage{xcolor}
\usepackage[T1]{fontenc}
\usepackage{multirow}
\usepackage{booktabs}
\usepackage{hyperref}
\usepackage{bbm} 
\usepackage{amssymb}
\usepackage{wrapfig} 
\usepackage{array}
\usepackage{bm}
\usepackage{amsmath}
\usepackage[misc]{ifsym}
\usepackage{graphicx,verbatim}
\begin{document}
\title{EndoGen: Conditional Autoregressive Endoscopic Video Generation}
%

\author{Xinyu Liu$^{1}$, Hengyu Liu$^{1}$, Cheng Wang$^{1}$, Tianming Liu$^{2}$, Yixuan Yuan$^{1, \textrm{\Letter}}$}  
 
\authorrunning{Liu et al.}
\institute{$^1$ The Chinese University of Hong Kong, Hong Kong SAR \\
$^2$ University of Georgia, GA, USA\\
    \email{yxyuan@ee.cuhk.edu.hk}}

\maketitle              
\begin{abstract}
Endoscopic video generation is crucial for advancing medical imaging and enhancing diagnostic capabilities. However, prior efforts in this field have either focused on static images, lacking the dynamic context required for practical applications, or have relied on unconditional generation that fails to provide meaningful references for clinicians. Therefore, in this paper, we propose the first conditional endoscopic video generation framework, namely EndoGen. Specifically, we build an autoregressive model with a tailored Spatiotemporal Grid-Frame Patterning (SGP) strategy. It reformulates the learning of generating multiple frames as a grid-based image generation pattern, which effectively capitalizes the inherent global dependency modeling capabilities of autoregressive architectures. Furthermore, we propose a Semantic-Aware Token Masking (SAT) mechanism, which enhances the model's ability to produce rich and diverse content by selectively focusing on semantically meaningful regions during the generation process. Through extensive experiments, we demonstrate the effectiveness of our framework in generating high-quality, conditionally guided endoscopic content, and improves the performance of downstream task of polyp segmentation. Code released at \url{https://www.github.com/CUHK-AIM-Group/EndoGen}.


\end{abstract}
\begin{figure}[t]
    \centering
    \includegraphics[width=\textwidth]{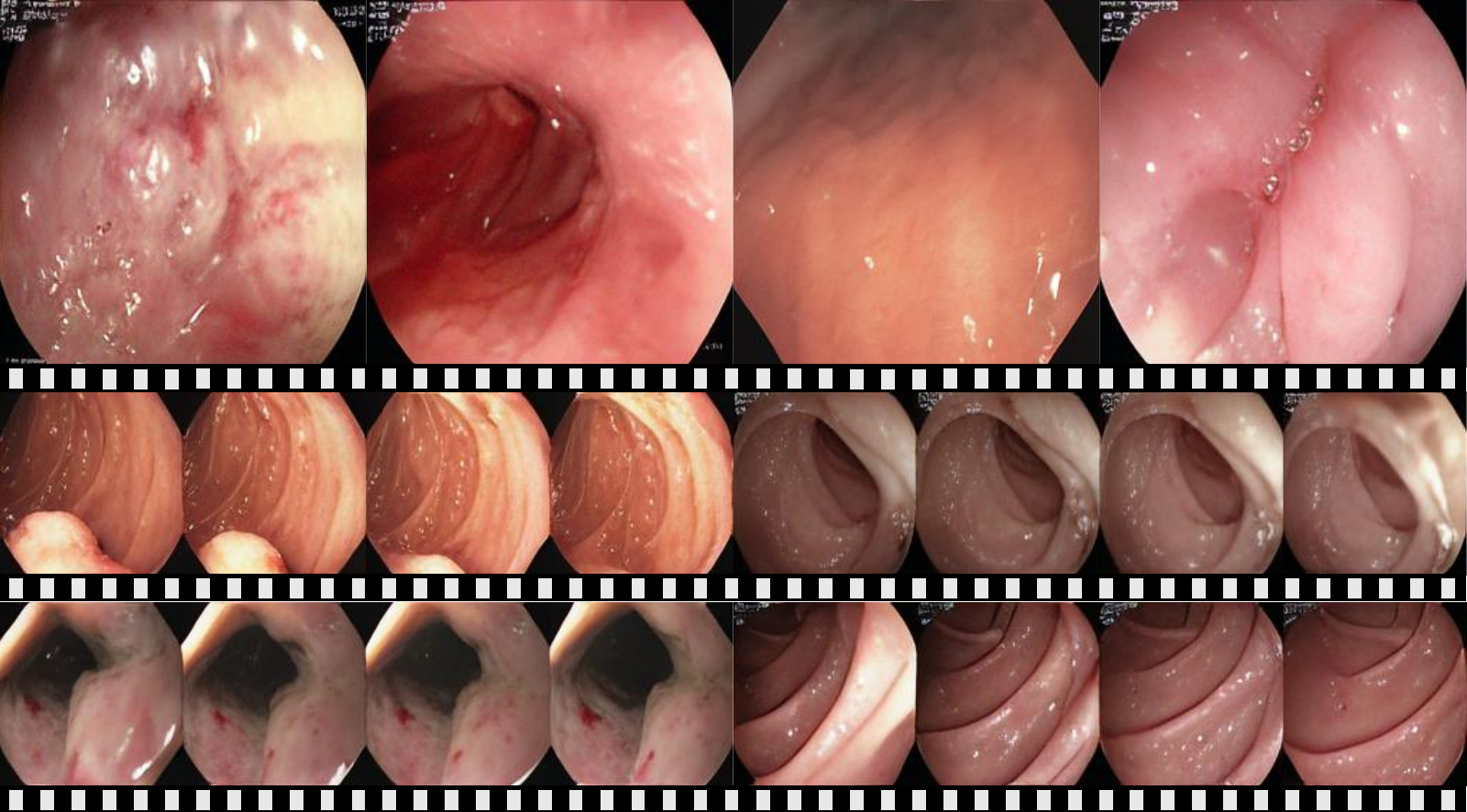}
    \caption{Endoscopic frames and videos with different resolutions generated by EndoGen.}
    \label{fig:teaser}
\end{figure}

\section{Introduction}

Endoscopy video generation is a critical task with far-reaching implications for medical applications, including surgical training, diagnostic system development, and patient education \cite{wang2023foundation, li2024endora, liu2022source, li2024static}. Realistic and controllable video synthesis can simulate rare pathological conditions, enable personalized surgical planning, and provide high-quality datasets for training AI models. However, existing generation methods primarily focus on static image synthesis \cite{sharma2024controlpolypnet, diamantis2022endovae} or unconditional video generation \cite{li2024endora}. Static images lack the temporal dynamics essential for simulating endoscopic procedures \cite{singer2022make}. For unconditional video models \cite{li2024endora}, they produce arbitrary sequences that are not aligned with specific anatomical or pathological conditions when needed by doctors \cite{yellapragada2024pathldm}. These limitations hinder their practical utility in scenarios requiring targeted outputs, such as generating videos of specific pathologies or tailoring simulations for surgical training. Thus, there is an urgent need for a conditional endoscopy video generation framework that can produce high-quality videos tailored to specific anatomical or procedural constraints.

Recent advances in autoregressive (AR) models \cite{touvron2023llama, lee2022autoregressive} have demonstrated superior conditional modeling capabilities compared to diffusion-based methods, particularly in tasks requiring long-range dependencies, such as text and image generation \cite{tian2025var, sun2024llamagen, li2025mar, yan2021videogpt}. 
With a condition token, AR models operate by predicting the next token based on all previously generated tokens, enabling them to capture complex hierarchical relationships and generate highly coherent outputs. 
However, despite their strengths, AR models are typically data-hungry \cite{deng2024nova, touvron2023llama} and have been largely confined to static image generation. Extending these models to endoscopy video generation poses significant challenges, as naive approaches often result in temporal inconsistencies and fail to leverage the inherent long-range dependencies of video data \cite{yan2021videogpt}. This raises an important question: 
\textit{Can we adapt the long-range conditional modeling capabilities of AR models to generate temporally coherent and contextually relevant endoscopic videos?}

To address this challenge, we initially construct a framework for conditional endoscopic video generation, named EndoGen. 
Specially, we develop a Spatiotemporal Grid-Frame Patterning (SGP) strategy to effectively train the AR model to learn spatial and temporal dependencies simultaneously. SGP redefines multi-frame generation as a synthesis task of a grid of interconnected images, which leverages the inherent capability of AR in modeling long-range relationships while preserving inter-frame continuity. 
This approach allows the generation of temporally consistent and detail preserved endoscopic sequences. Furthermore, to enhance the diversity and clinical relevance of the generated videos, we introduce a Semantic-Aware Token Masking (SAT) mechanism. 
SAT dynamically masks video tokens with less or redundant information, while preserving those with rich semantic content based on their intrinsic feature variance. This design encourages the model to focus on informative features that align closely with clinical objectives. With the proposed learning strategies, our framework is capable to generate highly realistic endoscopic videos across various conditions. We display generated frames and videos with different resolutions in Fig. \ref{fig:teaser}.

We extensively evaluate our method on video generation and downstream task. Experimental results demonstrate that EndoGen generates temporally coherent and clinically relevant endoscopic videos, outperforming existing methods in terms of both visual fidelity and utility for downstream application. Our work not only advances the state of the art in medical video generation but also opens new avenues for leveraging AR models in dynamic medical imaging tasks.

\section{Methodology}

\begin{figure}[t]
    \centering
    \includegraphics[width=0.99\textwidth]{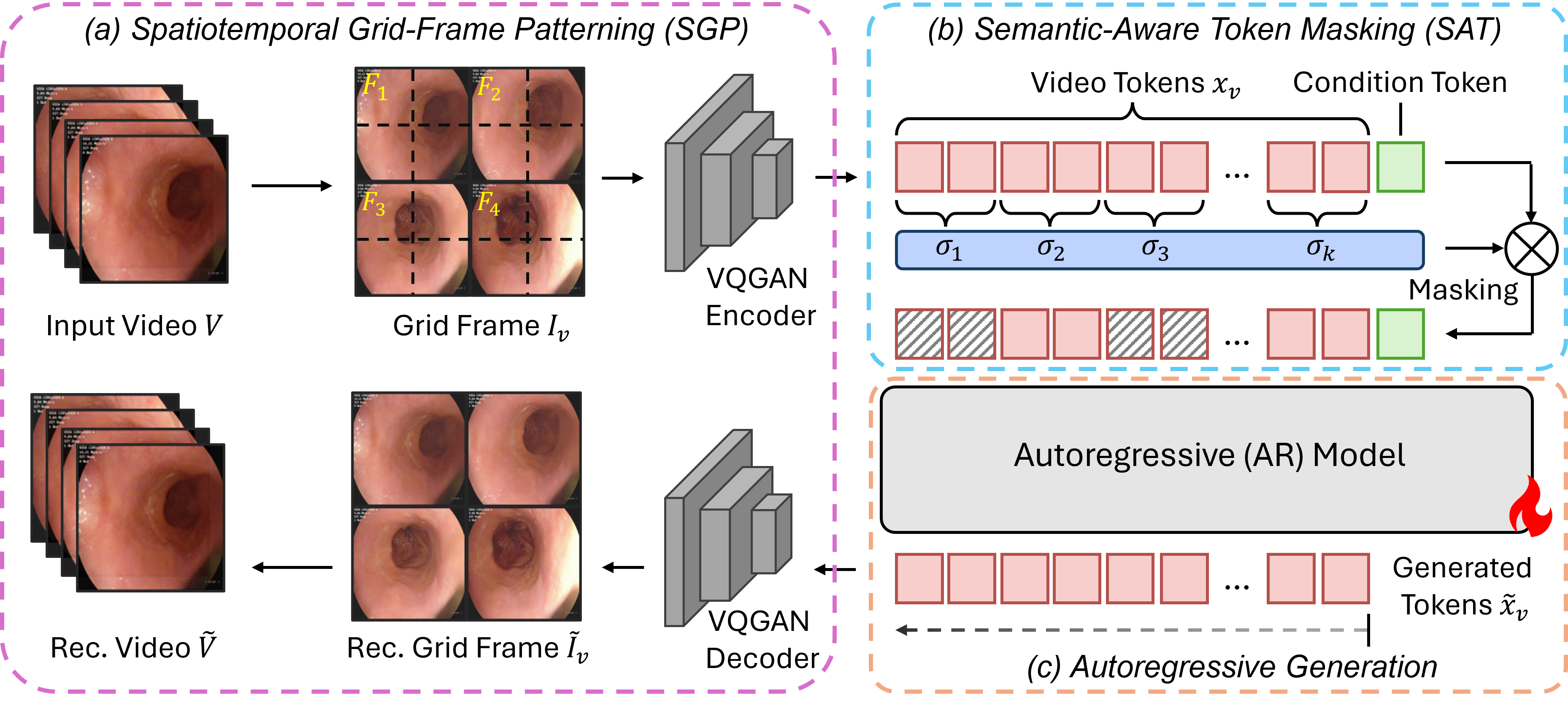}
    \caption{Illustration of the EndoGen framework. During training, each input video undergoes (a) Spatiotemporal Grid-Frame Patterning (SGP), (b) Semantic-Aware Token Masking (SAT), and (c) Autoregressive Generation. During inference, video tokens are generated autoregressively based on the provided condition token and then reconstructed into video format.}
    \label{fig:main_endogen_v2}
\end{figure}

The overview of EndoGen is presented in Fig. \ref{fig:main_endogen_v2}. During training, it reformulates the input video as grid frames with SGP (Sec. \ref{endogen:sgp}) and generates video tokens. Then, the video tokens are adaptively masked with SAT (Sec. \ref{endogen:lat}) to learn more diverse content. Specially, a conditional token is indexed from a set of learnable embeddings \cite{sun2024llamagen}, and serves as the starting prefilling token. Starting from it, the model generates a sequence of video tokens autoregressively. Without conditional token, the model can only generate random class samples and fails to produce desired class videos when needed by doctors. After concatenating the masked video and condition tokens, we feed them into the AR model to generate tokens autoregressively, and a standard cross-entropy loss \cite{tian2025var} is utilized for supervision of the generated token. At inference time, only a condition token is provided to the AR model and the generated tokens are decoded and reconstructed into the original video format.

\subsection{Spatiotemporal Grid-Frame Patterning (SGP)}
\label{endogen:sgp}

To bridge the gap between text/image and video generation in autoregressive models, we propose SGP, an effective strategy to encode both spatial and temporal information into a unified representation, which is shown in Fig. \ref{fig:main_endogen_v2}(a). Traditional video generation approaches either process videos with 3D blocks \cite{ho2022videodiffusionmodels} or with interleaved spatial and temporal modules \cite{ma2024latte, chen2024videocrafter2}. However, the 3D block-based methods suffer from high computational complexity and memory requirements during training \cite{mittal2021surveyon3dconv}, while the interleaved spatial-temporal methods introduce architectural complexity and could struggle in maintaining temporal consistency \cite{yan2019stat}. 

Different from them, our method maps the temporal sequence into a spatial representation, enabling simultaneous modeling of spatial and temporal dependencies via attention computation. Specifically, for each input video sequence $V$ with frames $\{F_1, F_2, ..., F_N\}$, we arrange them in a specific grid-based pattern $I_v$, and the $I_v$ is fed into a VQGAN \cite{esser2021taming} encoder to obtain the latent feature $x_v = E(I_v)$. Specifically, SGP arranges video frames in a sequential, row-by-row format within a large image, which ensures the frames maintain temporal dynamics when processed by the AR model. Afterwards, $x_v$ is processed with the proposed SAT (described in Sec. \ref{endogen:lat}) and reconstructed with the AR model in an autoregressive manner. The reconstructed latent representation $\tilde{x}_v$ is subsequently processed through the VQGAN decoder, which generates a reconstructed grid frame pattern $\tilde{I}_v=D(\tilde{x}_v)$. Finally, the framework decomposes the $\tilde{I}_v$ back into individual frames and reorders them to form the output video sequence $\tilde{V}$, ensuring temporal coherence throughout the generation process. SGP efficiently compresses temporal information into spatial patterns while preserving frame-to-frame relationships, which ensures the consistency of generation results.

\subsection{Semantic-Aware Token Masking (SAT)  }
\label{endogen:lat}

To enhance the diversity and clinical relevance of the generated videos, we introduce a SAT mechanism, as shown in Fig. \ref{fig:main_endogen_v2}(b). SAT dynamically prioritizes tokens with rich semantic content based on their intrinsic feature variance during training, while masking those with less informative or redundant features. This selective masking operation ensures that the model focuses on capturing informative features that are more aligned with clinical objectives, such as lesion areas or surgical tools.

Specifically, we are given a tokenized video feature $x_v$ of shape $(B,T \times L,C)$, where $B$ is the batch size, $T$ is the number of frames, $L$ is the token length of a single frame, and $C$ is the feature dimension. Specially, we first split the feature with $(T \times L)/H$ segments, with each has a token length of $H$. For each segment, the variance across the channel dimension is computed, and a masking ratio is adaptively determined based on the variance:
\begin{equation}
\sigma^2_i = \frac{1}{H} \sum_{h=1}^{H} (s_{i,h} - \mu_i)^2, \quad 
p_{\text{i}} = \text{Clamp}\left(\left(1 - \frac{\sigma^2_i}{\max(\sigma^2_i)}\right) \cdot p_{\text{max}}, 0, p_{\text{max}}\right),
\end{equation}
where $\mu_i$ and $\sigma^2_i$ are the mean and variance values for the $i$-th segment $s_{i}$, and $p_{\text{max}}$ is the maximum threshold for the masking ratio. During training, a binary mask $M_{i}$ is applied to each segment based on the computed ratio, ensuring that only the most informative tokens are retained:
\begin{equation}
s'_{i} = s_{i} \odot M_{i}, \quad \text{where} \quad M_{i} \sim \text{Bernoulli}(1 - p_{\text{i}}).
\end{equation}
With SAT, the model is encouraged to generate videos that are not only temporally coherent but also semantic meaningful, addressing a critical limitation of existing video generation methods.

\section{Experiments}

\subsection{Datasets and Implementation Details} 

We conduct experiments on two endoscopic video datasets. HyperKvasir \cite{borgli2020hyperkvasir} contains videos with 8 different pathological findings: \{barretts, cancer, esophagitis, gastric-antral-vascular-ectasia, gastric-banding-perforated, polyps, ulcer, varices\}. SurgVisdom \cite{zia2021surgvisdom} contains surgical videos on porcine model with 3 surgical tasks: \{dissection, knot-tying, needle-driving\}. The AR model is trained for 300 epoch using AdamW optimizer with learning rate 1e-4. $H$ is set to 8. In the main comparison experiments, we use 16-frame video clips from the datasets with a specific sampling interval, and resize each frame to the 128$\times$128 resolution for training. We also present results for videos with 64 frames or a spatial resolution of $256\times256$ in the supplementary material. 
We apply a frozen VQGAN pretrained on general domain data \cite{esser2021taming} to reduce training cost, and use the ImageNet pretrained class conditional image generation model \cite{sun2024llamagen} as the AR model weight initialization. We compare with diffusion based methods SimDA \cite{xing2024simda} and VDM \cite{ho2022videodiffusionmodels}, as well as the autoregressive method VideoGPT \cite{yan2021videogpt}. 
Per-class Fréchet Video Distance (FVD) \cite{fvd}, Content-Debiased Fréchet Video Distance (CD-FVD) \cite{cdfvd}, Fréchet Inception Distance (FID) \cite{fid}, and Learned Perceptual Image Patch Similarity (LPIPS) \cite{lpips} are used as the evaluation metrics. For all these metrics, lower values indicate better performance.

\subsection{Video Generation Performance}

\begin{table}[t]
\caption{Conditional video generation FVD results on HyperKvasir \cite{borgli2020hyperkvasir} with different pathological findings, where lower values are better. Bold denotes best performance. \label{table:main_endogen}}
\centering
\resizebox{\textwidth}{!}{\setlength{\tabcolsep}{1.1mm}{\begin{tabular}{l|cccccccc|c}
\toprule
{Method} & {Bar.} & {Cancer} & {Eso.} & {Ecta.} & {Perf.} & {Polyps} & {Ulcer} & {Varices} & {{Avg.}} \\
\midrule
 SimDA \cite{xing2024simda} & 3479.1 & 5065.1 & 2041.4 & 3643.6 & 1641.7 & 3656.2 & 3688.0 & 3919.3 & 3391.8 \\
  VDM \cite{ho2022videodiffusionmodels} & 1758.8 & 4635.1 & 1366.9 & 2057.6 & 897.0 & 2348.3 & 2172.6 & 1766.5 & 2125.4 \\
  VidGPT \cite{yan2021videogpt} & 1433.1 & 2965.7 & 955.7 & 1649.4 & 636.3 & 1705.1 & 1616.1 & 1427.7 & 1548.6 \\
  \midrule
  \textbf{EndoGen} & 402.1	&908.0	&286.3	&628.1	&300.6	&423.2	&496.6	&612.9&\textbf{507.2}
\\
\bottomrule
\end{tabular}}}
\end{table}

\begin{table}[t]
\begin{minipage}{0.46\linewidth}
\centering
\caption{Conditional video generation FVD results on SurgVisdom \cite{zia2021surgvisdom} with different surgical tasks, lower is better.}
    \begin{tabular}{l|ccc|c}
        \toprule
        Method         & Dis. & Knot. & Dri. & Avg. \\ \midrule
        SimDA \cite{xing2024simda}         & 3682.8     &  5889.2      & 3342.7     & 4304.9    \\
        VDM \cite{ho2022videodiffusionmodels}          & 1948.2 &  2716.3    &  2365.4      & 2343.3        \\
        VidGPT \cite{yan2021videogpt}      &  3394.5    &  2397.5      &  2197.6    &   2663.2  \\
        \midrule
        \textbf{EndoGen} & 1324.9     & 1606.5       & 1249.5     & \textbf{1393.6}    \\ \bottomrule
    \end{tabular}\label{tab:results_sv}
\end{minipage}
\hspace{2em}
\begin{minipage}{0.46\linewidth}
\centering
\caption{Results comparison on the HyperKvasir \cite{borgli2020hyperkvasir} dataset with different evaluation metrics, lower values are better.}
\begin{tabular}{l|cccc}
        \toprule
        Method         & CD-FVD & FID & LPIPS  \\ \midrule
        SimDA \cite{xing2024simda}   & 1319.9  & 288.4 &  0.565    \\
        VDM \cite{ho2022videodiffusionmodels}   & 851.4        &  246.8 & 0.652 \\
        VidGPT \cite{yan2021videogpt}      & 980.6  & 235.8   &  0.563 \\
        \midrule
        \textbf{EndoGen} & \textbf{765.3}  & \textbf{76.56}   &  \textbf{0.528}      &  \\ \bottomrule
    \end{tabular}\label{tab:different_metrics_endogen}
\end{minipage}
\end{table}

\begin{table}[htbp]
    \centering
    \setlength{\tabcolsep}{4.5pt}
    \caption{Ablation of the components and maximum threshold for the masking ratio on the HyperKvasir dataset. Lower values denote better performance.}
    \begin{tabular}[width=\textwidth]{l|cc|cccc}
        \toprule
        Metric & w/o SGP & w/o SAT & $p_\text{max}$=0.1 & $p_\text{max}$=0.2 & $p_\text{max}$=0.3 & $p_\text{max}$=0.4 \\ \midrule
        FVD    & 2617.5      & 562.0       & 533.8
         & 514.8        & 507.2      &   521.2      \\ \bottomrule
    \end{tabular}
    \label{tab:endogen_ablation}
\end{table}

\begin{figure}[h]
\hspace*{-0.5cm}
    \centering
    \includegraphics[width=0.90\textwidth]{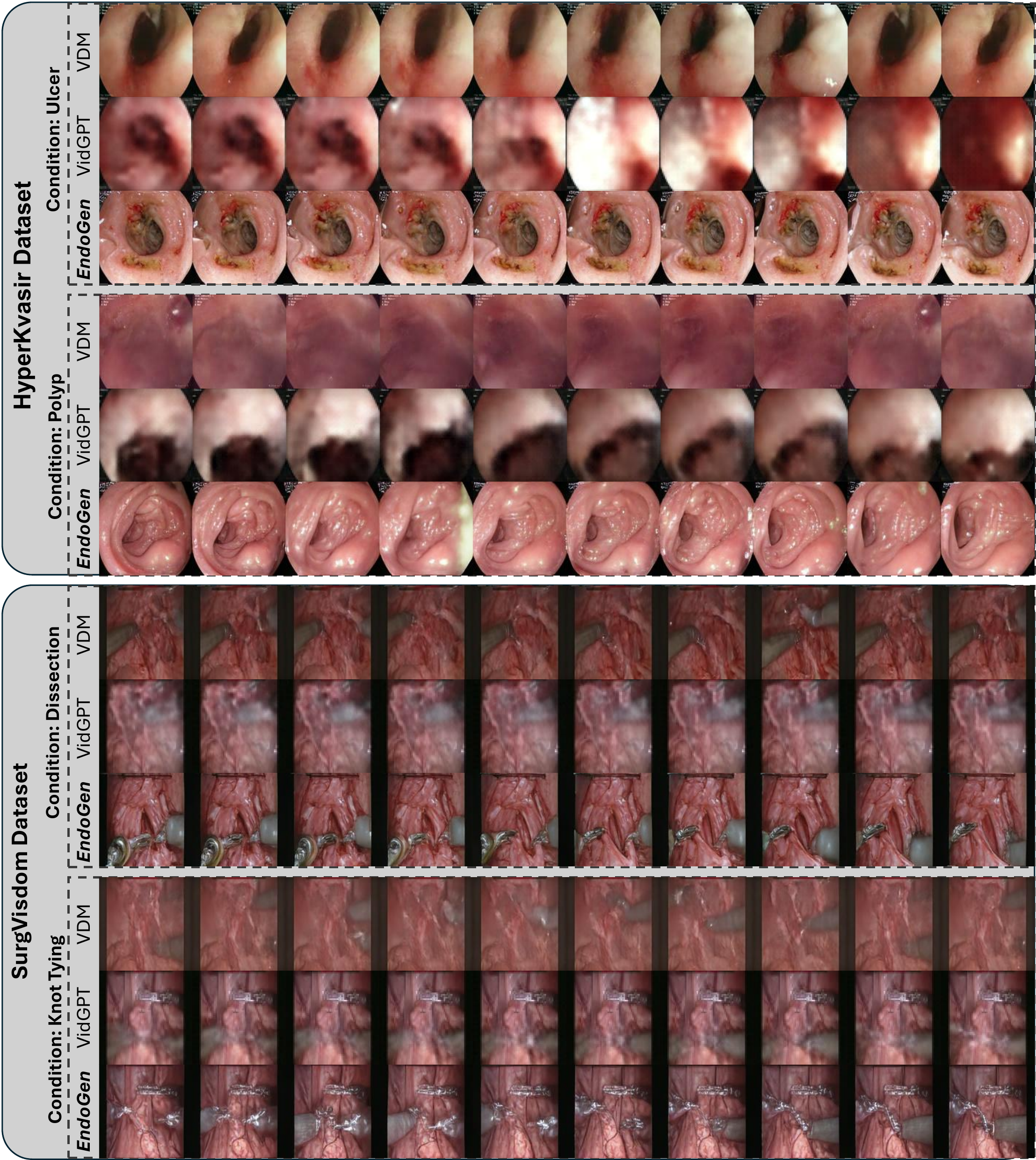}
    \caption{Qualitative comparison on the HyperKvasir \cite{borgli2020hyperkvasir} and SurgVisdom \cite{zia2021surgvisdom} datasets with different conditions.}
    \label{fig:endogen_dataset1}
\vspace{-5pt}
\end{figure}

\textbf{Comparison with State-of-the-arts.} As shown in Table \ref{table:main_endogen}, EndoGen achieves state-of-the-art performance across all eight pathological findings in the conditional generation on HyperKvasir, outperforming existing methods by significant margins. It is observed that diffusion-based approaches like VDM \cite{ho2022videodiffusionmodels} could struggles with fine-grained anatomical consistency (e.g., 2172.6 FVD for ulcers), while our method shows a significantly reduced FVD value of 496.6. Compared to the autoregressive models like VidGPT \cite{yan2021videogpt}, EndoGen demonstrates a better ability in generating complex pathologies such as varices, with 612.9 vs 1427.7 FVD. Notably, EndoGen shows particular strength in capturing subtle variations in Barrett's esophagus, which is attributed to our SAT mechanism that prioritizes diagnostically relevant features. In the qualitative comparison in Fig. \ref{fig:endogen_dataset1}, EndoGen demonstrates more anatomically accurate and temporally coherent endoscopic videos under different conditions. Meanwhile, the generated videos show clearer textures and smoother transitions.

In Table \ref{tab:results_sv}, EndoGen also shows superior performance on the SurgVisdom dataset, achieving 40.6\% lower FVD compared to the diffusion-based VDM. This demonstrates its robustness to diverse procedural dynamics. From Fig. \ref{fig:endogen_dataset1}, compared to other methods \cite{ho2022videodiffusionmodels, yan2021videogpt} that show distorted or blurry content in the challenging task, EndoGen offers superior visual representation of tissues and equipment, meanwhile effectively capturing the characteristics of the corresponding surgical task. Furthermore, Table \ref{tab:different_metrics_endogen} reveals that EndoGen achieves state-of-the-art results across various key metrics. 
These results validate that EndoGen could effectively leverage the long-range dependency modeling of autoregressive models in diverse scenarios in endoscopic video generation.

\noindent\textbf{Ablation Studies.} In Table \ref{tab:endogen_ablation}, we ablate the components of EndoGen on the HyperKvasir dataset. Replacing SGP with a simple 2D reshaping of the video sequence results in a significant decline in performance, demonstrating the effectiveness of the proposed grid frame in capturing spatial and temporal information. Removing SAT also leads to reduced diversity and fidelity in the generated videos. Additionally, we explored various maximum thresholds $p_\text{max}$ for masking, and setting it to 0.3 yields optimal performance, striking a balance between model learning complexity and capability enhancement.

\subsection{Downstream Task: Semi-supervised Polyp Segmentation}

\begin{table}[t]
\caption{\label{table:downstream_polypseg}Performance comparison on the semi-supervised polyp segmentation task. Blue subscript denotes the improvement over the supervised baseline. The fg and bg denotes the foreground polyp and the background regions, respectively. Lab. denotes labeled real data. Unl.-Real denotes unlabeled real data. Unl.-Syn denotes unlabeled synthetic data by EndoGen. Bold refers to the best result.}
    \centering
    \setlength{\tabcolsep}{3.5pt}
    \begin{tabular}{l|ccc|lll}
        \toprule
        Method & {Lab.} & {Unl.-Real} & {Unl.-Syn} & {Dice (\%)} & {IoU$_{fg}$ (\%)} & {IoU$_{bg}$ (\%)} \\ \midrule
        {Supervised} & \checkmark & & & 69.75 & 61.72 & 90.58 \\
        \midrule
        \multirow{3}{*}{{FixMatch \cite{sohn2020fixmatch}}} & \checkmark & \checkmark & & 70.80 & 62.66 & 91.19 \\
        & \checkmark & & \checkmark & $70.96_{\textcolor{blue}{\uparrow 1.21}}$ & $62.79_{\textcolor{blue}{\uparrow 1.07}}$ & $91.49_{\textcolor{blue}{\uparrow 0.91}}$ \\
        & \checkmark & \checkmark & \checkmark & $\textbf{71.03}_{\textcolor{blue}{\uparrow 1.28}}$ & $\textbf{63.14}_{\textcolor{blue}{\uparrow 1.42}}$ & $\textbf{91.66}_{\textcolor{blue}{\uparrow 1.10}}$ \\
        \midrule
        \multirow{3}{*}{{PolypMix \cite{jia2024polypmixnet}}} & \checkmark & \checkmark & & 87.13 & 82.59 & 95.55 \\
        & \checkmark & & \checkmark & $87.84_{\textcolor{blue}{\uparrow 18.09}}$ & $82.40_{\textcolor{blue}{\uparrow 20.68}}$ & $95.47_{\textcolor{blue}{\uparrow 4.89}}$  \\
        & \checkmark & \checkmark & \checkmark & $\textbf{87.92}_{\textcolor{blue}{\uparrow 18.17}}$ & $\mathbf{82.41}_{\textcolor{blue}{\uparrow 20.69}}$ & $\textbf{95.77}_{\textcolor{blue}{\uparrow 5.19}}$  \\
        \bottomrule
    \end{tabular}
\end{table}

\begin{figure}[t]
    \centering
    \includegraphics[width=\textwidth]{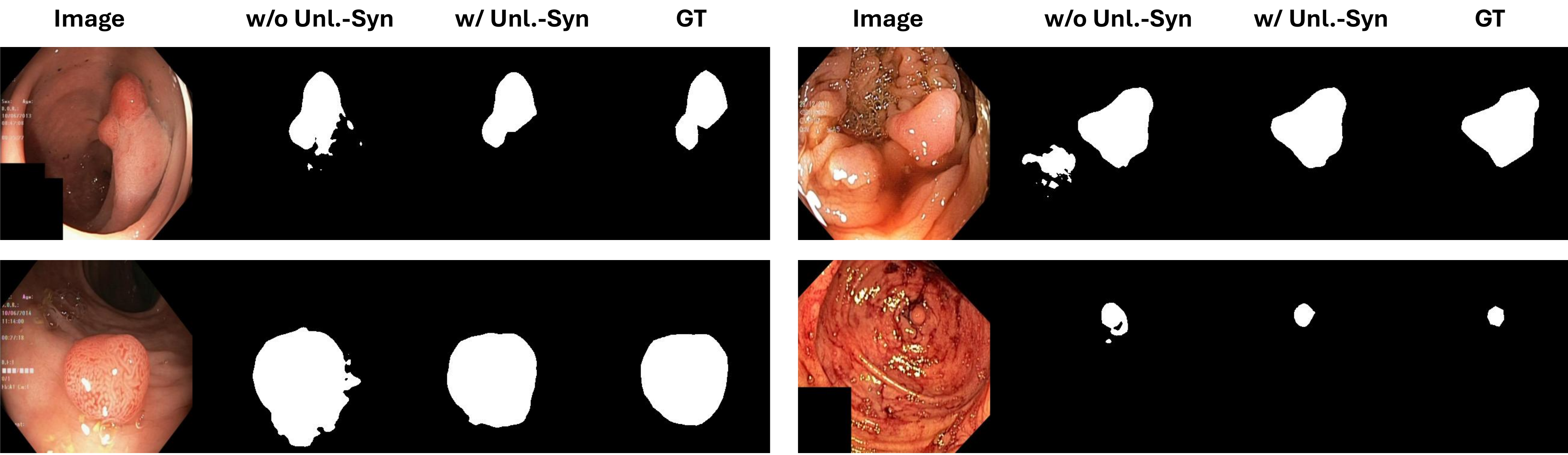}
    \caption{Qualitative results of semi-supervised polyp segmentation.}
    \label{fig:endogen_downstream}
\end{figure}

Semi-supervised medical image segmentation is an essential approach that reduces the labeling cost for improved performance \cite{liu2024diffrect}. 
To evaluate the fidelity of EndoGen synthetic videos, we generate polyp frames as the unlabeled data for the semi-supervised polyp segmentation task, and train the segmentation model different semi-supervised methods \cite{sohn2020fixmatch, jia2024polypmixnet}. We compare three training settings: using real unlabeled data (Unl.-Real); using synthetic unlabeled data (Unl.-Syn); and using both real and synthetic unlabeled data.
We utilize 1,000 images from the HyperKvasir polyp segmentation dataset \cite{borgli2020hyperkvasir}, splitting it into an 8:2 train-test ratio. In the training set, 10\% of the images are labeled, while the remaining are unlabeled. Additionally, we randomly sample the same number of synthetic frames generated by EndoGen to create the unlabeled synthetic set.
According to the results in Tab. \ref{table:downstream_polypseg}, replacing real with synthetic data could even yield higher Dice scores of 70.96\% with FixMatch \cite{sohn2020fixmatch} and 87.84\% with PolypMix \cite{jia2024polypmixnet}, demonstrating that EndoGen-generated data could effectively serves as a substitution of real data. Moreover, combining real and synthetic further improves performance, which indicates that the synthetic data complements real data and enhances overall segmentation quality. Fig. \ref{fig:endogen_downstream} gives a qualitative comparison between the segmentation results without and with our synthetic data. From the two cases in the left column, segmented results with our synthetic data capture better polyp structure and demonstrate more accurate boundary. In the right column, model trained with our additional unlabeled data performs better on small objects and effectively reduces false positives.

\section{Conclusion}

In this paper, we introduce EndoGen, an innovative framework for conditional autoregressive endoscopic video generation. EndoGen reformulates video sequence learning as a grid-frame pattern using SGP, and we propose an SAT strategy to enhance the diversity and clinical relevance of the generated results. Extensive validation has shown its superiority in both generation performance and downstream application. We hope that EndoGen will effectively support clinicians and advance research in medical generative models.

\begin{credits}
\subsubsection{\ackname} This work was supported by Innovation and Technology Commission Innovation and Technology Fund ITS/229/22, PRP/082/24FX and Hong Kong Research Grants Council (RGC) General Research Fund 14204321.

\subsubsection{\discintname}
The authors have no competing interests to declare that are
relevant to the content of this article.
\end{credits}
%

%
%
\bibliographystyle{splncs04}
\bibliography{Paper-1015}
%




\end{document}